# Unsupervised Measure of Word Similarity: How to Outperform Co-occurrence and Vector Cosine in VSMs

Enrico Santus[*], Tin-Shing Chiu[*], Qin Lu[*], Alessandro Lenci[§], Chu-Ren Huang[*]

[*] The Hong Kong Polytechnic University, Hong Kong
esantus@gmail.com, cstschiu@comp.polyu.edu.hk, {qin.lu, churen.huang}@polyu.edu.hk

[§] University of Pisa, Italy
alessandro.lenci@ling.unipi.it

**Abstract**

In this paper, we claim that *vector cosine* – which is generally considered among the most efficient unsupervised measures for identifying word similarity in *Vector Space Models* – can be outperformed by an unsupervised measure that calculates the extent of the intersection among the most mutually dependent contexts of the target words. To prove it, we describe and evaluate *APSyn*, a variant of the *Average Precision* that, without any optimization, outperforms the *vector cosine* and the *co-occurrence* on the standard ESL test set, with an improvement ranging between +9.00% and +17.98%, depending on the number of chosen top contexts.

## Introduction and Related Work

Word similarity detection plays an important role in Natural Language Processing (NLP), as it is the backbone of several applications, such as paraphrasing, query expansion, word sense disambiguation, automatic thesauri creation, and so on (Terra and Clarke, 2003).

Several approaches have been proposed to measure word similarity (Jarmasz and Szpakowicz, 2003; Levy et al., 2015). Some of them are lexicon-based, while others are corpora-based. The latter generally rely on the *distributional hypothesis*, according to which words that occur in similar contexts also have similar meanings (Harris, 1954). Although all of them extract statistics from large corpora, they vary in the way they define what has to be considered context (i.e. lexemes, syntax, etc.) and in which way such context is used (Santus et al., 2014a; Hearst, 1992).

A common way to represent word meaning in NLP is by using vectors to encode the Strength of Association (SoA) between the target word and its contexts. In the resulting Vector Space Model (VSM), *vector cosine* is then generally used to calculate word similarity by measuring the distance between these vectors (Turney and Pantel, 2010).

A well-known problem with the statistical approaches is that they rely on a very loose definition of similarity. Indeed, according to the *distributional hypothesis*, similarity does not only include synonyms, but also other semantic relations, such as hypernymy, co-hyponymy and even antonymy (Santus et al., 2014b-c). For this reason, several datasets have been proposed by the NLP community to test distributional similarity measures (Santus et al., 2015). One of the most used is the *English as a Second Language* dataset (ESL), introduced in Turney (2001). It consists of 50 multiple-choice synonym questions, with 4 choices each.

In this paper, we describe and evaluate *APSyn*, a completely unsupervised measure that calculates the extent of the intersection among the $N$ most related contexts of two target words, weighting such intersection according to the rank of the contexts in a mutual dependency ranked list. In our experiments, *APSyn* outperforms the cosine, with an improvement ranging between +9.00% and +17.98% in the ESL test set, depending on the chosen $N$.

## Method and Evaluation

The *vector cosine*, described in the following equation (where $f_{xi}$ is the $i$-th feature in the vector $x$), calculates word similarity by looking at the normalized correlation between the contexts distribution of two words, $w_1$ and $w_2$:

$$\cos(w_1, w_2) = \frac{\sum_{i=1}^{n} f_{1i} \times f_{2i}}{\sqrt{\sum (f_{1i})^2} \times \sqrt{\sum (f_{2i})^2}}$$

Our hypothesis is that similar words share more mutually dependent contexts than less similar ones. A way to test it is: i) measuring the intersection among the $N$ most relat-ed contexts of two target words, and ii) weighting such intersection according to the rank of the shared contexts in the dependency ranked lists. For every target word, in fact, we rank all contexts according to the *Local Mutual Information* values (LMI; Evert, 2005) and pick the top $N$:

$$APSyn(w_1, w_2) = \sum_{f \in N(F_1) \cap N(F_2)} \frac{1}{(rank_1(f_1) + rank_2(f_2))/2}$$

That is, for every feature $f$ included in the intersection between the top $N$ features of $w_1$, $N(F_1)$, and $w_2$, $N(F_2)$, *APSyn* will add 1 divided by the average rank of the feature, among the top LMI ranked features of $w_1$, $rank_1(f)$, and $w_2$, $rank_2(f)$.

### Evaluation

**VSM**. We use a window-based VSM recording word co-occurrences within the 5 nearest content words to the left and right of each target. Co-occurrences are extracted from a combination of ukWaC and WaCkypedia corpora (around 2.7 billion words) and weighted with LMI.

**TEST SET**. For evaluation, we use the ESL dataset, introduced in Turney (2001) as a way of evaluating algorithms for measuring the degree of word similarity. The test set consists of 50 multiple-choice synonym questions, with 4 choices. Each question was turned into four pairs, having as first word the *problem word* and as second word one of the possible choices. Some words have been lemmatized, in order to have a correspondent form in the VSM.

**TASK**. We have assigned *APSyn* scores to all the pairs, and then sorted them in a decreasing order. We considered positive every problem word having the correct answer on top, negative the others. 5 out of 50 questions were excluded, because the correct answers were not present in the VSM. 6 out of 45 questions had one wrong choice missing. In this case, only 0.75 was added if the answer was correct.

**RESULTS**. In Table 1, we report the results in the test.

| Measure | N | Full + | 75% + | Correct |
| --- | --- | --- | --- | --- |
| **APSyn** | 100 | 24 | 3 | **26.25**/45 = **58.33%** |
| **APSyn** | 1000 | 22 | 3 | **24.25**/45 = 53.89% |
| **Cosine** | --- | 20 | 3 | **22.25**/45 = 49.44% |
| **Co-occ.** | --- | 14 | 4 | **17.00**/45 = 37.78% |

*Table 1. Correct answers in the ESL test set for APSyn (N=1000, 100), vector cosine and co-occurrence.*

### Discussion and Conclusions

In this paper, we have described *APSyn*, a measure based on the calculation of the shared top ranked features, and its performance on the ESL questions. The measure outperforms the *vector cosine* and the *co-occurrence*, plus several lexicon-based and hybrid models[1]. Even though the performance needs to be improved by optimizing the model (i.e. learning the value of $N$ from a training set), our experiments show that the intersection among the $N$ most related contexts of the target words is in fact an index of similarity. It is relevant to mention here also the role of $N$: the larger the amount of considered contexts, the lower the ability of identifying similarity. This also confirms our hypothesis that similar words share a significant number of top mutually dependent contexts, but such intersection becomes less significant when not only the top contexts are considered. Since ESL is small, the reported results should be further investigated on a larger dataset (Santus et al., 2015).[2]

### References

Evert, S. 2005. *The Statistics of Word Cooccurrences: Word Pairs and Collocations*. Dissertation, University of Stuttgart.

Harris, Z. 1954. Distributional structure. *Word*, Vol. 10 (23). 146-162.

Hearst, M. A. 1992. Automatic Acquisition of Hyponyms from Large Text Corpora. *Proceedings of the 14th International Conference on Computational Linguistics*. 539-545.

Jarmasz, M. and Szpakowicz, S. 2003. Roget's thesaurus and semantic similarity, *Proceedings of RANLP 2003*. 212-219.

Levy, O., Goldberg, Y. and Dagan I. 2015. Improving Distributional Similarity with Lessons Learned from Word Embeddings. *TACL 2015*.

Santus, E., Lenci, A., Lu, Q., and Schulte im Walde, S.. 2014a. Chasing Hypernyms in Vector Spaces with Entropy. *Proceedings of EACL 2014*, 2:38–42.

Santus, E., Lu, Q., Lenci, A. and Huang, C-R. 2014b. Unsupervised Antonym-Synonym Discrimination in Vector Space. *Proceedings of CLIC-IT 2014*.

Santus, E., Lu, Q., Lenci, A. and Huang, C-R. 2014c. Taking Antonymy Mask off in Vector Space. *Proceedings of PACLIC 2014*.

Santus, E., Yung, F., Lenci, A. and Huang, C-R. 2015. EVALution 1.0: an Evolving Semantic Dataset for Training and Evaluation of Distributional Semantic Models. *Proceedings of ACL-IJCNLP 2015*, 64.

Terra, E. and Clarke, C.L.A. 2003. Frequency estimates for statistical word similarity measures. *Proceedings of HLT/NAACL 2003*. 244–251.

Turney, P.D. and Pantel, P. 2010. From Frequency to Meaning: Vector Space Models of Semantics. *Journal of Articial Intelligence Research*, Vol. 37. 141-188.

Turney, P.D. 2001. Mining the Web for synonyms: PMI-IR versus LSA on TOEFL. *Proceedings of ECML-2001*. 491-502.

---

[1] http://aclweb.org/aclwiki/index.php?title=ESL_Synonym_Questions_(State_of_the_art)

[2] More info: https://github.com/esantus/. This work is partially supported by HK PhD Fellowship Scheme under PF12-13656.